\documentclass[10pt,letterpaper]{article}
\usepackage[margin=1in,footskip=0.5in]{geometry}

\usepackage{amsmath,amssymb}
\usepackage{changepage}
\usepackage{textcomp,marvosym}
\usepackage{cite}
\usepackage[right]{lineno}
\usepackage[nopatch=eqnum]{microtype}
\DisableLigatures[f]{encoding = *, family = * }
\usepackage[table]{xcolor}
\usepackage{array}
\usepackage{lipsum}
\usepackage{authblk}
\usepackage{longtable}
\usepackage{booktabs}
\usepackage{lastpage,fancyhdr,graphicx}
\usepackage{epstopdf}
\usepackage{nameref,hyperref}

\newcolumntype{+}{!{\vrule width 2pt}}
\newlength\savedwidth

\usepackage[aboveskip=1pt,labelfont=bf,labelsep=period,singlelinecheck=off]{caption}

\makeatletter
\renewcommand{\@biblabel}[1]{\quad#1.}
\makeatother

\title{Crop Pest Classification Using Deep Learning Techniques: A Review}

\author[1]{Muhammad Hassam Ejaz}
\author[1,*]{Muhammad Bilal}
\author[1]{Usman Habib}
\author[2]{Muhammad Attique}
\author[2,*]{Tae-Sun Chung}

\affil[1]{FAST School of Computing, National University of Computer and Emerging Sciences, Islamabad, Pakistan}
\affil[2]{Department of Artificial Intelligence, Ajou University, Suwon-Si, South Korea}
\affil[*]{Corresponding authors: muhammad.bilal@isb.nu.edu.pk; tschung@ajou.ac.kr}

\date{}

\begin{document}
\maketitle

\begin{abstract}
Insect pests continue to pose a serious threat to crop yields around the world, and traditional methods for monitoring them are often slow, manual, and difficult to scale. In recent years, deep learning has emerged as a powerful solution, with techniques like convolutional neural networks (CNNs), vision transformers (ViTs), and hybrid models gaining popularity for automating pest detection. This review looks at 37 carefully selected studies published between 2018 and 2025, all focused on AI-based pest classification. The selected research is organized by crop type, pest species, model architecture, dataset usage, and key technical challenges. The early studies relied heavily on CNNs, but the latest work is shifting toward hybrid and transformer-based models that deliver higher accuracy and better contextual understanding. Still, challenges like imbalanced datasets, difficulty in detecting small pests, limited generalizability, and deployment on edge devices remain significant hurdles. Overall, this review offers a structured overview of the field, highlights useful datasets, and outlines the key challenges and future directions for AI-based pest monitoring systems.
\end{abstract}


\section{Introduction}
\label{sec:introduction}
Insect pests are one of the most persistent challenges in modern agriculture, and global estimates indicate that up to 40\% of food crops are lost annually to plant pests and diseases \cite{FAO2021}. Traditional pest monitoring, relying on manual inspections by farmers and entomologists, is labor-intensive, time-consuming, subjective, and impractical for large-scale agriculture \cite{teixeira2023systematic}. Recent advancements in deep learning, notably convolutional neural networks (CNNs) and vision transformers (ViTs), have shown great promise, with a recent tomato study reporting 90\% training accuracy, 87\% validation accuracy, and 85\% testing accuracy on an augmented eight-category insect-pest dataset \cite{alshannaq}. However, applying these technologies in real-world agricultural settings introduces complexities like variable lighting, occlusions, and complex backgrounds, significantly impacting performance.

Pest identification in agricultural fields has evolved rapidly from basic CNN architectures to sophisticated hybrid models that combine CNNs with vision transformers, leveraging the strengths of both local feature extraction and global context interpretation \cite{utku2025convvit}. Recent innovations incorporating attention mechanisms and multi-scale processing have improved detection of small, camouflaged pests in complex environments \cite{wang1122}. Additionally, optimized mobile implementations now enable real-time pest identification via smartphone cameras, enhancing accessibility for farmers directly in the field \cite{berka2023cactivit}. Despite these technological advancements, dataset limitations persist, including severe class imbalance and insufficient representation of rare or region-specific pests, limiting model generalizability \cite{wu2019ip102,teixeira2023systematic}. Emerging methods such as quantum-inspired CNNs \cite{amin2} and diffusion-based data augmentation \cite{du2025longtail} show potential but require further validation under practical agricultural conditions.

This review aims to provide a comprehensive analysis of research progress in deep learning-based crop pest classification, focusing on different types of crops and pests. The paper has four key contributions. First, we present a taxonomy of pest classification approaches in Section \ref{sec:taxonomy}, organizing studies by pest types, target crops, techniques, datasets, and challenges. The studies based on different crops and pests are discussed in Sections~\ref{sec:crops_addressed} and \ref{sec:pests}, respectively. Second, we provide detailed insights on technical approaches, including CNN-based methods (Section \ref{sec:cnn_methods}), vision transformers (Section \ref{sec:vision_transformers}), hybrid architectures (Section \ref{sec:hybrid_architectures}), and object detection models (Section \ref{sec:object_detection}). Third, we analyze benchmark datasets and their characteristics in Section \ref{sec:datasets}, identifying gaps in current data resources. Finally, we discuss important research gaps and open challenges that future studies need to address in Section \ref{sec:challenges_future} and present the conclusion in Section \ref{sec:practical_applications}.

\section{Materials and Methods}
\label{sec:materials_methods}

\begin{figure*}[!ht]
    \centering
    \includegraphics[width=\textwidth]{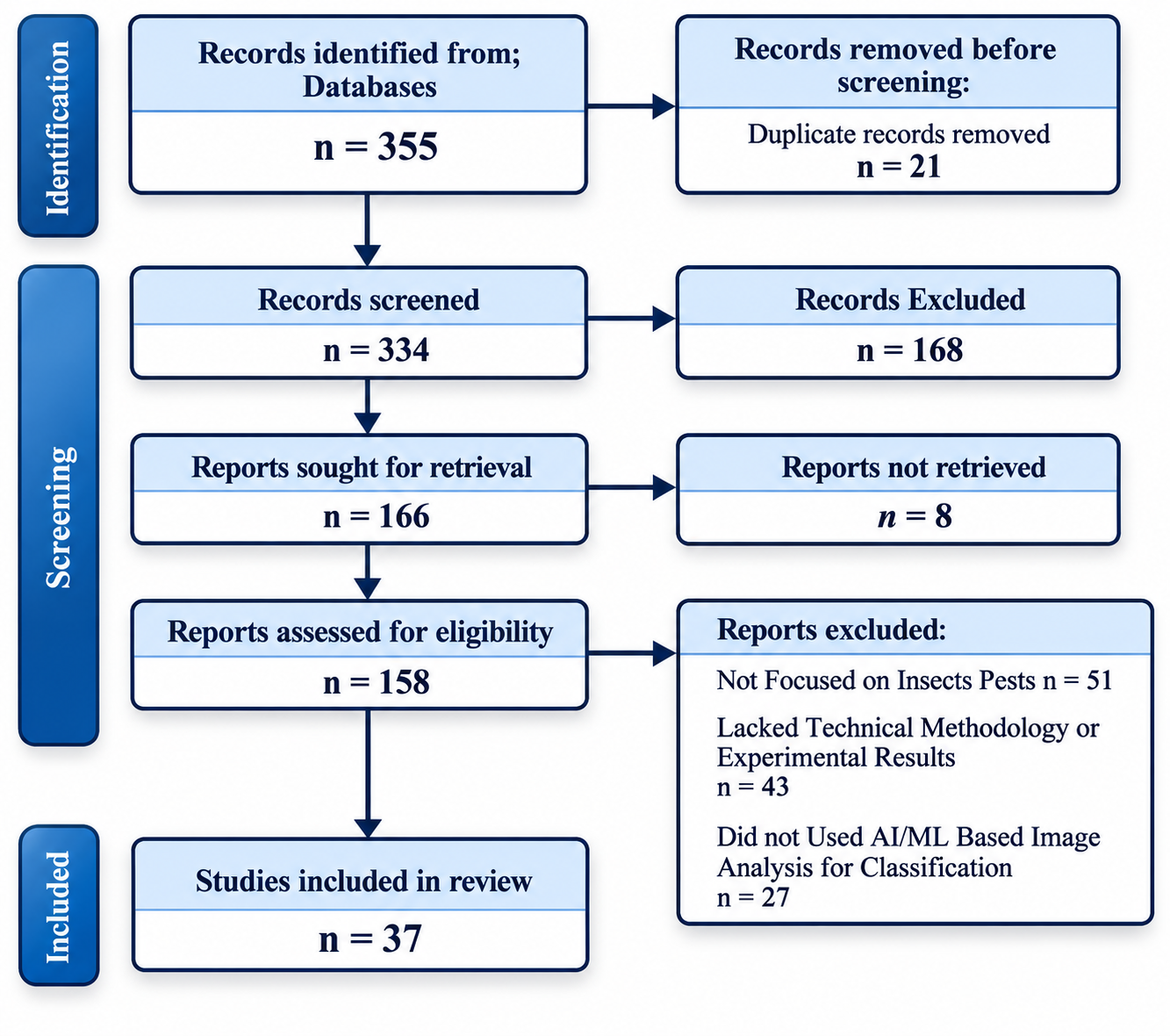}
    \caption{PRISMA flow diagram for literature selection.}
    \label{fig:prisma}
\end{figure*}

To conduct this review, a structured and systematic search strategy was utilized to identify relevant studies on insect pest classification employing deep learning and artificial intelligence (AI) techniques. The search covered publications from approximately 2018 to 2025, reflecting the rapid advancements in AI-driven agricultural applications. Multiple scholarly databases, including IEEE Xplore, ACM Digital Library, SpringerLink, ScienceDirect, and Google Scholar, were searched using keywords such as ``insect pest detection", ``pest classification", ``deep learning of pests of crops", ``plant disease and pest identification", ``YOLO pest detection," and ``vision transformer agriculture." Initial searches identified 355 articles. After removing duplicate entries (n = 21), 334 records remained. These records underwent a screening process based on their titles and abstracts, which led to the exclusion of 168 irrelevant articles, leaving 166 for further review. Subsequently, full-text retrieval was conducted for these 166 articles. Eight reports could not be retrieved. The inclusion criteria applied during full-text assessment required that (1) studies focused explicitly on insect pests affecting crops (excluding studies exclusively addressing plant diseases or weeds), (2) methods involved AI or machine learning with computer vision, (3) publications were peer-reviewed journal or conference papers or reputable preprints, and (4) studies provided adequate technical details and experimental results for evaluation. 

After the full-text assessment, 121 reports were excluded: 51 studies were not focused specifically on insect pests, 43 lacked detailed technical methodology or sufficient experimental results, and 27 did not employ AI-based or ML-based image analysis for classification. This rigorous selection process resulted in a final set of 37 studies included in this review. Throughout the process, efforts were made to ensure comprehensive coverage, including foundational early work in convolutional neural networks (CNNs) and recent advancements such as transformer models. Additionally, reference lists of key selected articles were reviewed to include significant studies potentially overlooked during the database search. The final collection of papers encompasses a global body of work, including several surveys that contextualize the field (e.g., Teixeira et al.\cite{teixeira2023systematic}) as well as diverse case studies on pest detection in different crops and environments. This systematic approach to literature gathering ensures that the insights drawn in this review are representative of the current state of research in AI-based pest classification. Figure \ref{fig:prisma} illustrates the process used for identifying, screening, and including studies in this review.

\section{Taxonomy of AI-Based Pest Classification}
\label{sec:taxonomy}

\begin{figure*}[!ht]
    \centering
    \includegraphics[width=\textwidth]{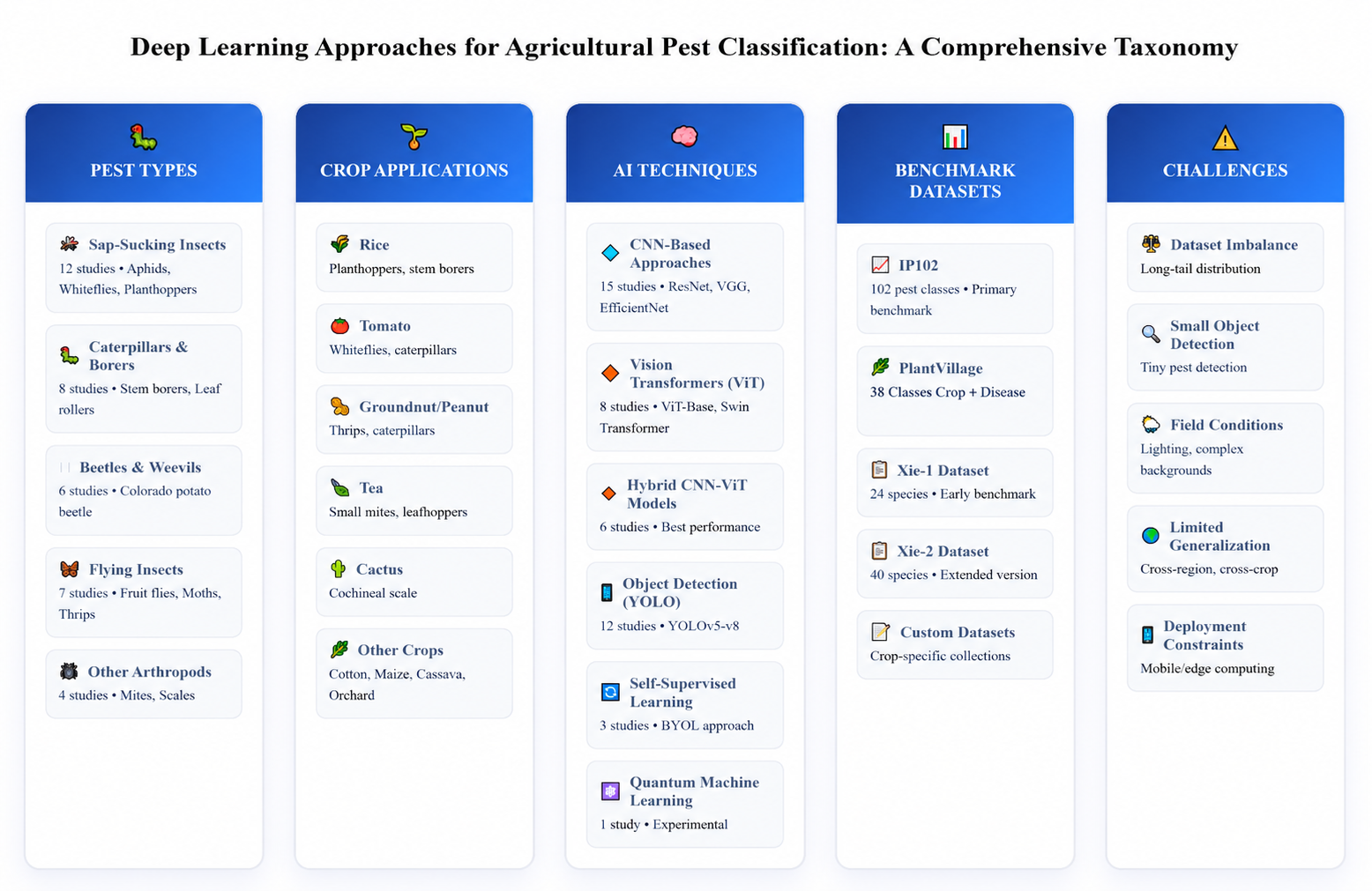}
    \caption{Taxonomy of pest classification approaches.}
    \label{fig:taxonomy}
\end{figure*}

This section presents a comprehensive taxonomy developed from the reviewed literature on AI-assisted pest classification. The taxonomy is organized around five key dimensions: the types of insect pests studied, the crops targeted, the AI techniques employed, the challenges encountered, and the datasets used (Figure \ref{fig:taxonomy}). This structured classification allows us to analyze trends across research works and highlight underexplored combinations, such as underrepresented pest species or limited use of hybrid attention-based models like HPMA-ViT. The following subsections present each of these dimensions with supporting literature, tables, and insights. This categorization provides a structured view of the research landscape.

\textbf{Scope and terminology:} The studies discussed in this review address several related but distinct tasks, and we use the following terms: \emph{Pest classification} assigns an image (or image region) to one of several pest categories and includes binary pest-presence classification and image-level classification of infestation severity. \emph{Pest detection} localizes individual pests in an image, typically with one-stage or two-stage object detectors. \emph{Pest counting} estimates the number or density of pests from images or detections. \emph{Disease recognition} identifies plant diseases from leaf images. Following the inclusion criteria stated in the Materials and Methods section, studies of insect pests are the primary focus. Studies of plant diseases and general computer-vision methods \cite{mohanty2,Ramcharan2017,chen2024using,dosovitskiy2021an} are discussed as methodological background or as analogies rather than as direct evidence about insect pests. 

\section{Crops Addressed}
\label{sec:crops_addressed}
Research on AI-driven pest classification has spanned a variety of crops, often focusing on those of high economic importance or with well-documented pest issues. We organize the literature by the main crops addressed, as different crops present different pest challenges and imaging conditions. Table \ref{tab:ai-crop} summarizes representative studies for each major crop, including the reference, year, brief description, and dataset used.

\begin{table*}[!ht]
\centering
\caption{\textbf{Summary of deep learning techniques for crop pest/disease detection.}
}
\tiny
\renewcommand{\arraystretch}{2}
\begin{tabular}{p{0.5cm} p{1.5cm} p{2.5cm} p{3cm} p{2.5cm} p{4cm} p{5.5cm}}
\hline
\textbf{Ref} & \textbf{Crop} & \textbf{AI Technique} & \textbf{Dataset (Images, Classes)} & \textbf{Results} & \textbf{Notes/Limits} \\[1.2ex]
\hline
\cite{alshannaq} & Tomato & CNN (custom + augmentation) & Tomato insect-pest images, 8 categories (5,894 after augmentation) & 85\% test accuracy (90\% training, 87\% validation) & Small original dataset ($\sim$842 images), augmentation essential, mobile use only. \\
\hline
\cite{berka2023cactivit} & Cactus & Vision Transformer (ViT‑B16) & CactiViT dataset (3921 images, 3 classes) & 88.7\% accuracy & Mobile real‑time app but dataset is small and public. \\
\hline
\cite{wu2019ip102} & Various (8 crops) & CNN (ResNet-50 baseline) & IP102 (75k images, 102 pests) & 49.4\% accuracy (baseline) & Large, imbalanced dataset. Fine-grained classes difficult. \\
\hline
\cite{amin2} & Rice & YOLOv5 + Quantum CNN (hybrid) & Paddy Pest (Kaggle: ~648 images, binary) & 99.56\% accuracy (binary) & Quantum approach but only binary classification (pest vs no pest). \\
\hline
\cite{mohanty2} & Multiple (14 crops) & CNN (AlexNet, GoogLeNet) & PlantVillage (54k images, 26 diseases) & 99.3\% accuracy (lab) & High accuracy in lab but not tested in field. \\
\hline
\cite{Ramcharan2017} & Cassava & CNN (ResNet-50, transfer) & 5,660 field images (5 classes) & 93\% accuracy (field) & Partial images give lower accuracies. \\
\hline
\cite{liuv3} & Tomato & Object Detection (YOLOv3+) & Tomato pest/disease images (custom) & 92.39\% accuracy & Improved multi‑scale detection but needs field validation. \\
\hline
\cite{li2022plant} & Multi-crop & Vision Transformer (ViT) & 39-class plant disease and insect pest images & 96.7\% accuracy & Outperformed CNNs but requires field testing. \\
\hline
\cite{Venkatasaichandrakanth} & Peanut & Transformer-based (EViTA, dual-branch ViT + MFO) & Peanut pests: aphids, wireworm, gram caterpillar (3 datasets) & $\sim$92\% accuracy for EViTA+PCA+MFO & Only 3 pests with limited number of samples \\
\hline
\cite{venkatasaichandrakanth2024gnvit} & Peanut & Vision Transformer (GNViT) & IP102 peanut-pest subset (4,157 images, 4 classes) & 99.52\% training accuracy, precision/recall/F1 $\sim$0.90-0.92 & Pure ViT with a small and imbalanced dataset. \\
\hline
\cite{xiao2024ga} & Multi-crop & Efficient CNN (GA‑GhostNet) & IP102 (102 classes) and others via transfer & 71.9\% accuracy (IP102), 95\% on simpler tasks & Lightweight model (~4M parameters) and struggled on 102‑class task. \\
\hline
\cite{Du2024} & Corn \& Rice & Object Detection (YOLOv7) & Rice-corn pest dataset & 93.3\% mAP (mean Average Precision) on the rice-corn pest set & Optimized for small pests and outperforms the YOLOv7 baseline by 2.1 points. \\
\hline
\cite{wang2025insect} & General & Object Detection (YOLOv8) & Insect‑YOLO (3M parameters, custom data) & 93.8\% mAP (field) & Tiny model for edge devices but effective on low‑resolution images. \\
\hline
\cite{Kar} & Multi-crop & Self-supervised (SSL ResNet) & Unlabeled + labeled pest images & 93-94\% accuracy with fewer labels & SSL reduced label needs, but it is not widely adopted. \\
\hline
\end{tabular}
\begin{flushleft}
\footnotesize
\textit{Note:} Results are quoted as reported by each study and are not directly comparable. Classification accuracy and detection mAP (mean Average Precision) measure different things. mAP values may use different IoU thresholds and evaluation protocols, and training, validation, and test results are distinguished where reported.
\end{flushleft}
\label{tab:ai-crop}
\end{table*}

\subsection{Rice}
\label{subsec:rice}
Rice is a staple crop worldwide and hosts numerous insect pests (planthoppers, stem borers, leaf folders, etc.). Several studies have targeted rice pest identification. For example, Liu et al.\cite{liu2016paddy} developed a CNN-based system that combined saliency-map localization with a deep CNN to classify 12 common paddy field insect pests, using a dataset of over 5,000 images covering leafhoppers, brown planthoppers, and other rice pests. Their approach achieved promising accuracy, although the dataset was modest in size compared with later benchmarks such as IP102. More recently, object detection models have been applied to rice pests. Du et al.\cite{Du2024} introduced an improved YOLOv7 model (named YOLOv7-PSAFP) to detect small pests in rice and corn fields. By enhancing the feature pyramid and loss functions to better capture tiny objects, their model reached 93.3\% mAP on a rice-corn pest dataset, 2.1 percentage points above the standard YOLOv7 baseline on the same data. Another example is the work by Javeria Amin et al.\cite{amin2}, who combined YOLOv5 for rice pest localization with a quantum convolutional network for classification. They treated it as a binary problem (paddy with pest vs. without pest) and reported extremely high accuracy (99.56\%) on their paddy field images. These rice-focused studies highlight how both CNN classifiers and one-stage detectors have been customized for the rice domain. However, multi-class rice pest identification (distinguishing among many pest species in rice) remains challenging unless large labeled datasets are available. The IP102 benchmark, which includes many rice pest species, has exposed the difficulty of fine-grained classification with a baseline accuracy below 50\% (ResNet-50) \cite{wu2019ip102}. Nonetheless, progress is steadily being made via improved models and data augmentation targeted at rice pest images \cite{wu2022attention}.

\subsection{Tomato}
\label{subsec:tomato}

Tomato crops also feature prominently in pest classification research due to their economic value and susceptibility to pests (whiteflies, caterpillars, etc.). Al-Shannaq et al.\cite{alshannaq} focused on identifying insect infestations on tomatoes using a CNN combined with extensive image augmentation. They assembled a small dataset of tomato insect-pest images in eight categories and expanded it to 5,894 images with augmentation techniques. The results for the CNN model reached 90\% training accuracy, 87\% validation accuracy, and 85\% testing accuracy across the eight categories. The authors suggest that the model could be used in mobile applications to assist farmers. Notably, without augmentation, the performance dropped to nearly 50-60\%, underscoring how limited data can hurt generalization. This study’s outcomes indicate that augmentation was critical to reaching high accuracy on the small tomato dataset, and it suggested the need for larger, more diverse training data. In another tomato-related work, Liu and Wang \cite{liuv3} applied the YOLOv3 object detection model (with multi-scale image pyramids) to detect tomato pests and diseases in images. Their improved YOLOv3 could identify multiple issues on tomato plants (like pest presence and disease symptoms) with a high detection accuracy of 92.39\% and a detection time of 20.39 ms. The authors also compared the method with Faster R-CNN and the original YOLOv3. Challenges noted in tomato pest studies include dealing with complex backgrounds (foliage, varying lighting) that can confuse classifiers. For instance, Ramcharan et al.\cite{Ramcharan2017} achieved 93\% accuracy on field images of cassava diseases but observed slightly lower classification accuracy in partial leaf images. Overall, the literature on tomatoes shows that CNNs and YOLO detectors can be effective for targeted pest problems, provided the models are trained on sufficiently representative images or enhanced via transfer learning and augmentation. Table \ref{tab:ai-crop} lists these tomato-focused studies, noting their datasets (often custom images from greenhouses or fields) and results.

\subsection{Groundnut}
\label{subsec:groundnut}

Peanut crops (groundnuts) have been the subject of pest classification research, especially in recent years with the advent of transformer models. Two studies by Venkatasaichandrakanth and Iyapparaja address peanut pest identification with transformer-based models. In the first, they proposed an Enhanced Vision Transformer Architecture (EViTA) that uses a double-layer transformer encoder to merge image segments of different sizes, with Moth Flame Optimization (MFO) used for feature selection, and evaluated it on three peanut pest datasets: aphids, wireworm, and gram caterpillar \cite{Venkatasaichandrakanth}. In a follow-up study, they introduced GNViT, a pure Vision Transformer (ViT) model trained on a four-class groundnut pest subset of IP102 (aphids, armyworms, wireworms, and thrips, 4,157 images), and reported 99.52\% training accuracy with validation accuracy of 95.31\% and precision, recall, and F1-score between 0.90 and 0.92 \cite{venkatasaichandrakanth2024gnvit}. They compared GNViT against LeNet, AlexNet, GoogLeNet, ResNet, and EViTA with PCA and MFO (accuracies of 56-72\% for the CNNs and 92\% for EViTA), although those baseline figures are carried over from earlier work. Taken together, these studies suggest that attention-based models can improve accuracy over traditional CNN-only baselines for the fine-grained task of identifying peanut pests. The datasets involved are small (GNViT's four classes are heavily imbalanced, with 2,456 aphid images against 527 for thrips), so results may be optimistic for those specific conditions. Nonetheless, the peanut case exemplifies how researchers are pushing beyond CNNs to boost pest classification performance in specific crops.

\subsection{Tea}
\label{subsec:tea}
Tea plantations face unique pest challenges, including small arthropods such as mites and insects such as leafhoppers that can be hard to spot. Recent AI research on tea crops has emphasized detecting tiny pests on tea leaves. For instance, Zejun Wang et al.\cite{Wang2024micro} proposed a lightweight YOLOv8-based model to detect microscopic pests in tea gardens. By introducing attention mechanisms and adaptive convolution modules, their improved YOLOv8 achieved significantly higher mAP than the baseline, increasing mAP by 3.12\%, 4.34\%, 5.44\%, 16.54\%, and 11.29\% in YOLOv10, YOLOv9, YOLOv7, Faster RCNN, and SSD models, respectively, enabling fast and accurate identification of early-stage tea pests. This indicates the model could reliably spot small sap-sucking pests (like tea green leafhoppers or mites) that often appear as only a few pixels in images. Earlier work by Samanta et al.\cite{samanta2012tea} had used traditional feature selection and neural networks on a much smaller tea pest dataset (8 species, 609 records) and reported a 100\% correct classification rate using ANN. Across these studies, tea emerges as a crop where high image resolution and specialized models are needed to overcome background clutter (leaf textures) and the minute size of pests. The improved results with attention-enhanced YOLO models (e.g., adding spatial attention to focus on pest regions) show that deep learning can indeed tackle even the small-pest problem in tea. Table \ref{tab:ai-crop} compiles the efforts in the tea domain, including the datasets (often custom-collected images of tea leaves in field conditions) and techniques ranging from classical to cutting-edge.

\subsection{Cactus}
\label{subsec:cactus}
An unusual but noteworthy case is pest management in cactus cultivation. Berka et al.\cite{berka2023cactivit} introduced CactiViT, a vision transformer-based system to detect infestations of the carmine cochineal (a scale insect) on prickly pear cacti. This pest is devastating in regions like Morocco and requires early intervention. The authors created a new image dataset of cochineal-infested cacti and developed a mobile application powered by a ViT model to classify infestation levels. Their transformer model (ViT-B16 backbone) achieved 88.7\% classification accuracy on the cactus infestation dataset, outperforming comparable CNN models by about 2.6\%. Impressively, the system runs in real time on smartphones, allowing farmers to photograph a cactus and immediately receive an assessment of pest presence and severity. This study demonstrates the viability of transformers even on relatively small, custom datasets, aided by transfer learning and the strong representational capacity of ViTs. It also highlights how AI can be applied beyond staple crops, tailored to specific regional pests. The CactiViT dataset has been made publicly available, which is valuable given the lack of open data on this pest. While cactus farming is a niche compared to rice or maize, the success of CactiViT underlines that AI-assisted pest management principles extend to any crop with a suitable dataset and model; even a challenging pest like cochineal (which blends into its host plant) can be reliably detected. This example pushes the envelope on deploying AI (transformers on mobile devices) for pest monitoring in the field.

\subsection{Other Crops} 
In addition to the above, many studies cover pests in other crops such as maize, cassava, cotton, and orchard trees. For instance, cassava pest and disease recognition was explored by Lin et al.\cite{lin2023GPAN} using a Graph Pyramid Attention Network. While they achieved near 99\% accuracy on a confined cassava disease dataset, the performance dropped to 57\% when tested on the diverse IP102 insect dataset. This suggests that models trained on one crop/pest scenario (cassava) may not generalize well to others without retraining. Maize (corn) pests like the fall armyworm have been targeted by object detectors. Researchers have developed optimized YOLO models for corn fields, as mentioned by Du et al.\cite{Du2024} for corn and rice pests. Cotton pests and soybean pests have likewise seen deep learning solutions, though they are fewer in number. Some recent works tackled orchard pests; for example, Liu et al.\cite{LIU2025109920} designed a transformer-based detector for forestry pests (e.g., insects on tree leaves) and reported 95.3\% mAP on a constrained forestry dataset, though this plummeted to 35.6\% mAP on the open-domain IP102 set. In summary, beyond the major crops, AI techniques have been applied to a wide range of agricultural contexts. However, success in one domain does not always translate to another, due to differences in pest appearance, background, and image capture conditions. Table \ref{tab:ai-crop} encompasses a selection of studies across various “other” crops, illustrating the breadth of applications and reminding us that each crop-pest combination may need specific consideration in model design and training.

\section{Pests}
\label{sec:pests}
\subsection{Sap-sucking Insects}
The insect pests addressed across the studies can be grouped into broad types. One major category is sap-sucking insects (order Hemiptera), which includes aphids, planthoppers, leafhoppers, whiteflies, and related pests. These are small insects that feed on plant sap and often occur in large numbers. They appear in multiple studies. Aphids are one of the target classes in peanut and groundnut pest works (\cite{Venkatasaichandrakanth}). They also appeared in the GA-GhostNet evaluation by Xiao et al.\cite{xiao2024ga}, which used the IP102 dataset. Planthoppers (e.g., the brown planthopper) are serious rice pests. Wang et al.\cite{wang2025insect} mention the rice planthopper and report improved precision for it using their Insect-YOLO model. Sap-sucking pests present challenges because of their small size, as noted by Wang et al.\cite{Wang2024micro}, baseline detectors had low precision on tiny planthoppers until attention mechanisms were added. These insects also often cause indirect damage by transmitting viruses, so identifying their presence early is critical.

\subsection{Caterpillars and Borers}
Another important group is caterpillars and borers, which are larvae of moths and butterflies (order Lepidoptera). These are chewing pests that consume leaves or bore into stems. Many studies include examples like Hu et al.\cite{hu2023detection} that dealt with rice pests like the rice leaf roller and stem borers (both are caterpillar stages of moths). The peanut “Gram Caterpillar” (likely referring to Helicoverpa or a similar genus) is included in Venkatasaichandrakanth and Iyapparaja \cite{Venkatasaichandrakanth}. Armyworms, a type of noctuid moth caterpillar, appear in the groundnut pest classification (Venkatasaichandrakanth and Iyapparaja \cite{venkatasaichandrakanth2024gnvit}) and in the IP102 dataset used by others. Caterpillars tend to create visible damage (holes in leaves, defoliation), which can aid detection; however, they can also hide (e.g., inside stems for borers), which is an ongoing challenge. Some models, like Liu et al.\cite{LIU2025109920}, aimed to improve detection of occluded or small pests, which could include larvae that are hidden or early instars that are tiny. The GA-GhostNet \cite{xiao2024ga} also mentions that their model was tested on the jute pest dataset and on an apple dataset, which includes jute hairy caterpillar and yellow mites.

\subsection{Beetles and Weevils}
Beetles and weevils (order Coleoptera) form another pest type covered in some datasets. For instance, the IP102 dataset (used in Lin et al.\cite{lin2023GPAN} and others) contains various beetle species (e.g., flea beetles, weevils). While specific beetle pests were not individually highlighted in many deep learning studies (perhaps due to fewer public image sets), the category remains important. For example, the Colorado potato beetle, various weevils in stored grains, etc., are potential targets for future AI models. One reason beetles may be less represented so far could be that some beetles are best detected with traps rather than imagery (e.g., pheromone traps for weevils).

\subsection{Other Insects}

Lastly, this category includes miscellaneous or less common pest types like flies, wasps, thrips, or any insect pests not in the above categories. Thrips (tiny, fringe-winged insects) were a class in the GNViT groundnut pest study \cite{venkatasaichandrakanth2024gnvit} (thrips often attack peanuts and soybeans by sucking plant juices similar to aphids). In general, the reviewed works collectively cover a wide taxonomic range from soft-bodied aphids to hard-shelled beetles and moth larvae, demonstrating the versatility of AI models. Each type poses its own detection challenges, e.g., aphids require high-resolution imagery to resolve their tiny bodies, moths/caterpillars require temporal monitoring since they grow quickly, and borers might require detecting indirect symptoms (like frass or entry holes) rather than the insect itself. Table \ref{tab:pest-ai-type} highlights these groupings, and indeed, many studies implicitly focus on one group or another (for example, the tea and cactus studies effectively dealt with pest damage symptoms, which are common for sap-suckers and disease-causing insects).

\begin{table}[!ht]
\centering
\caption{\textbf{Recent image-based AI studies for insect pest types and datasets.}}
\tiny
\renewcommand{\arraystretch}{2}
\begin{tabular}{p{2cm}@{\hspace{10pt}} p{2.3cm}@{\hspace{10pt}} p{4.7cm}@{\hspace{10pt}} p{6cm}}
\hline
\textbf{Ref} & \textbf{Pest Type} & \textbf{Examples (Species)} & \textbf{Datasets / Crops} \\
\hline
\cite{berka2023cactivit}, \cite{Wang2024micro}, \cite{anwar2022}, \cite{ung2021efficientinsectpestclassification} &
Sap‑sucking insects &
Aphids, whiteflies, leafhoppers, scales &
IP102, custom tea images, CactiViT (cactus) \\
\hline
\cite{Venkatasaichandrakanth}, \cite{venkatasaichandrakanth2024gnvit}, \cite{Du2024}, \cite{hu2023detection} &
Caterpillars \& Borers \&  Wireworm &
Rice leaf roller, Rice borers, gram caterpillar, armyworms &
IP102 (subsets), public peanut pest data, rice-corn pest images \\
\hline
\cite{dilex2023} &
Weevils &
Red Palm Weevil &
UAV Imagery \\
\hline
\cite{xing2020}, \cite{kumar2021} &
Other Pests &
Citrus pests, Locusts &
Lab Images, Locust Images \\
\hline
\end{tabular}
\label{tab:pest-ai-type}
\end{table}

\vspace{-2em}

\section{Techniques and Approaches}
\subsection{CNN-Based Approaches}\label{sec:cnn_methods}
Convolutional Neural Networks (CNNs) have been the workhorse of image classification and thus dominate early pest identification literature. These models excel at capturing local spatial features via convolutional filters. In pest detection, many studies leveraged classic CNN architectures pre-trained on ImageNet (e.g., VGG, ResNet) and then fine-tuned on pest datasets. For example, Ayan et al.\cite{ayan2020crop} tested multiple CNN architectures (VGG16, ResNet50, InceptionV3, Xception, MobileNet, etc.) on crop pest images, achieving 98.81\% accuracy on the D0 dataset. Similarly, Al-Shannaq et al.\cite{alshannaq} trained a CNN with extensive image augmentation to classify eight categories of tomato insect pests, reaching 85\% testing accuracy (90\% training, 87\% validation) and illustrating how CNNs combined with augmentation can perform reasonably well on small insect image datasets. To improve upon standard CNNs, researchers proposed specialized or optimized CNN variants. Sandhya Devi et al.\cite{devi2023efficientnetv2} employed an EfficientNetV2 model (a CNN with compound scaling) to handle the high variability in field-captured pest images, reporting improved accuracy of 80.1\% on the IP102 dataset for classifying crop pests. In another notable work, Xiao et al.\cite{xiao2024ga} introduced a custom lightweight CNN called GA-GhostNet, which integrates attention modules to drastically reduce model size without sacrificing accuracy. Their GhostNet-based classifier retained performance while being compact enough for deployment on resource-constrained devices. Wu et al.\cite{wu2022attention} embedded Efficient Channel Attention and Coordinate Attention modules in ResNet-50 for pest classification, improving accuracy by about 2\% with little extra computation. Wang et al.\cite{wang2023crop} proposed a three-scale CNN with attention (TSCNNA) for crop pest detection in the field, adding channel and spatial attention so that the network can focus on pests of different sizes against complicated backgrounds. They reported 93.16\% precision on a set of common crop pests, which they state is 3.7\% higher than VGG16. In a related study, Marinho et al.\cite{marinho2023automatic} applied CNNs to X-ray imagery of fruit fly pupae, achieving over 97\% accuracy in distinguishing parasitized versus healthy pupae, underscoring CNNs’ capability in even unconventional imaging modalities. Despite their effectiveness, CNNs have limitations when it comes to modeling global context or extremely fine-grained differences across an image. Pest species often require distinguishing subtle textures or shapes, and CNNs focusing on local receptive fields can miss these global patterns. These shortcomings of CNNs have paved the way for transformer-based architectures, which we discuss next (Dosovitskiy et al.\cite{dosovitskiy2021an}). Nonetheless, CNN-based solutions remain highly competitive for many classification tasks, often exceeding 90\% accuracy on curated pest image datasets, especially when sufficient training data is available and the model is carefully tuned. Table \ref{tab:cnn-litreview} presents the summary of the Key recent CNN-based approaches for crop pest and disease detection.

\begin{table}[!ht]
\centering
\caption{
\textbf{Literature Review Table on CNN-Based Approaches.}
}
\tiny
\renewcommand{\arraystretch}{2} 
\begin{tabular}{p{0.5cm} p{0.5cm} p{14cm}}
\hline
\textbf{Ref} & \textbf{Year} & \textbf{Description} \\[1.2ex]
\hline
\cite{alshannaq} & 2025 & Trained a CNN with extensive augmentation on an eight-category tomato insect-pest dataset (5,894 images after augmentation) reported 90\% training, 87\% validation and 85\% testing accuracy. \\[1.2ex]\hline
\cite{xiao2024ga} & 2024 & Proposed GA-GhostNet (a lightweight CNN with attention) achieving competitive accuracy with dramatically fewer parameters. \\[1.2ex]\hline
\cite{ayan2020crop} & 2020 & Evaluated several CNN architectures (e.g., VGG16, ResNet50) for multi‑pest classification, and the best model achieved 98.81\% on D0 dataset. \\[1.2ex]\hline
\cite{devi2023efficientnetv2} & 2023 & Employed EfficientNetV2 CNN to handle high variability in field images, improving pest classification accuracy up to 99.5\% on the Cassava dataset and 80.1\% on the IP102 dataset. \\[1.2ex]\hline
\cite{wang2023crop} & 2023 & Proposed TSCNNA, a three-scale CNN with channel and spatial attention for crop pest detection in the field, reported 93.16\% precision on a set of common crop pests. \\[1.2ex]\hline
\cite{marinho2023automatic} & 2023 & Applied CNN models to X‑ray images of fruit fly pupae, achieving over 97\% accuracy in distinguishing parasitized vs. healthy pupae. \\
\hline
\end{tabular}
\begin{flushleft}
\footnotesize
\end{flushleft}
\label{tab:cnn-litreview}
\end{table}

\subsection{Vision Transformers}\label{sec:vision_transformers}
Transformers, originally popularized in NLP, have emerged as powerful vision models due to their ability to capture global context via self-attention. Vision Transformers (ViTs), first introduced by Dosovitskiy and colleagues \cite{dosovitskiy2021an}, treat an image as a sequence of patches and learn long-range relationships, overcoming the locality bias of CNNs. Starting around 2020-2021, ViTs began to permeate pest classification. These models show particular value for high-resolution images or scenarios requiring holistic image understanding. In pest detection, Li et al.\cite{li2022plant} demonstrated that a ViT model, when sufficiently trained and augmented, could achieve up to 96\% accuracy on a mixed pest and plant disease dataset, outperforming comparable CNNs on the same data. This highlights transformers’ advantage in modeling global image features (such as an insect’s overall shape or the context of its surroundings), which are critical for distinguishing certain pest species. Similarly, Fu et al.\cite{fu2024} improved ViTs’ ability to focus on important regions by introducing a block-wise image partitioning before the self-attention layers. This modification effectively guided the transformer to attend to biologically significant areas (e.g., lesion spots or insect body parts) and yielded higher accuracies in crop disease identification. Venkatasaichandrakanth and Iyapparaja \cite{Venkatasaichandrakanth} proposed the Enhanced Vision Transformer Architecture (EViTA), a dual-branch ViT in which a double-layer transformer encoder merges image segments of different sizes and Moth Flame Optimization (MFO) is used for feature selection on peanut pest datasets. It was reported to perform better than or comparably to contemporary Vision Transformers and efficient CNNs. Saranya et al.\cite{SARANYA2024108584} modified the attention mechanism of the ViT itself. Their hybrid pooled multihead attention (HPMA) combines hybrid pooling with a modified attention mechanism so that both local and global features are captured, and they reported 98\% accuracy on a new 10-class pest dataset and 98\% and 95\% on two benchmark datasets, outperforming CNNs and standard ViTs. Transformers have also been integrated into object detection frameworks for pests. For example, Hu et al.\cite{hu2023detection} embedded a Swin Transformer module into a YOLO detector’s feature pyramid. The resulting model (sometimes termed “YOLO-GBS”) captured long-range dependencies in complex rice field images, improving pest detection performance under cluttered backgrounds. In another work, Liu et al.\cite{LIU2025109920} developed a transformer-based end-to-end detector for pests that adds feature compensation, local information enhancement and a regional pale-shaped self-attention (RPSA) mechanism to better detect small insects and distant objects. These studies demonstrate how transformers can overcome some CNN limitations, especially in global reasoning and nuanced differentiation among visually similar classes. It is worth noting that ViTs typically require large training datasets to realize their full potential, due to fewer built-in inductive biases than CNNs. In the agricultural domain, where data can be scarce, techniques like data augmentation and transfer learning are often employed alongside ViTs (Li et al.\cite{li2022plant}). 

Nonetheless, as open datasets grow and transformer architectures are refined, ViTs have begun to surpass CNNs in pest classification tasks that benefit from global feature modeling. The transformer approach marks a shift towards models that can consider an entire scene (e.g., a leaf and its pests) collectively rather than piecemeal. Table \ref{tab:vit-litreview} illustrates the key studies leveraging vision transformers for pest and disease detection in agricultural imagery.

\begin{table}[!ht]
\centering
\caption{
\textbf{Literature Review Table on Vision Transformer-Based Approaches.}
}
\tiny
\renewcommand{\arraystretch}{2}
\begin{tabular}{p{0.5cm} p{0.5cm} p{14cm}}
\hline
\textbf{Ref} & \textbf{Year} & \textbf{Description} \\[1.2ex]
\hline
\cite{dosovitskiy2021an} & 2021 & Introduced Vision Transformer (ViT) for image recognition, using self-attention on image patches instead of convolutions, a foundational work enabling transformers in vision. \\[1.2ex]\hline
\cite{li2022plant} & 2022 & Applied a ViT to pest images with extensive augmentation, achieving 96.7\% accuracy, outperforming CNN baselines on the same pest/disease dataset. \\[1.2ex]\hline
\cite{Venkatasaichandrakanth} & 2023 & Proposed EViTA, a dual-branch Vision Transformer whose double-layer encoder merges image segments of different sizes, with Moth Flame Optimization (MFO) used for feature selection, reported better or comparable performance versus contemporary ViTs and efficient CNNs on peanut pest datasets. \\[1.2ex]\hline
\cite{LIU2025109920} & 2025 & Developed a transformer-based end-to-end pest detector with feature compensation and regional pale-shaped self-attention (RPSA) that excelled at modeling global context and detecting small pests in images. \\[1.2ex]\hline
\cite{hu2023detection} & 2023 & Embedded a Swin Transformer into YOLO (YOLO-GBS) for rice pest detection, capturing long-range dependencies in cluttered field images, improving detection performance. \\[1.2ex]\hline
\cite{fu2024} & 2024 & Improved ViTs focus via block-wise image partitioning, helping the model attend to key pest regions (e.g., insect body), which boosted classification accuracy. \\[1.2ex]\hline
\cite{SARANYA2024108584} & 2024 & Proposed hybrid pooled multihead attention (HPMA) for a ViT, combining hybrid pooling with a modified attention mechanism to capture local and global features, reported 98\% accuracy on a new 10-class pest dataset and 98\% and 95\% on two benchmark datasets. \\[1.2ex]\hline
\end{tabular}
\begin{flushleft}
\footnotesize
\end{flushleft}
\label{tab:vit-litreview}
\end{table}

\subsection{Hybrid CNN-Transformer Architectures}\label{sec:hybrid_architectures}
To capitalize on both local and global feature learning, researchers have increasingly turned to hybrid models that combine CNN and transformer components. These architectures aim to harness CNNs’ prowess in capturing fine details (edges, textures) and transformers’ strength in modeling long-range relationships. A number of studies report that such hybrids outperform either pure CNNs or pure transformers in isolation. A hybrid design was used for tea pest detection by Hu et al.\cite{hu2024tea}. Their TP-YOLOX model combines convolution and multi-head self-attention in a single module and adds a multi-scale attention module, so that local insect features are integrated with global context such as the surrounding leaves. It improved mAP by 4.50\% over the original YOLOX-s on tea pests in complex backgrounds. Hybrids have also made their mark in object detection. Transformer-enhanced YOLO variants illustrate this trend. For instance, the aforementioned YOLO-GBS by Hu et al.\cite{hu2023detection}, which embeds a Swin Transformer module in a YOLO detector, improved pest detection in cluttered rice-field images, suggesting that mixing convolutional and attention mechanisms can improve the localization of pests in images. Recently, Utku et al.\cite{utku2025convvit} introduced ConvViT, a hybrid model for farm insect detection that combines a CNN feature extractor with a ViT classifier. ConvViT demonstrated strong performance on-device by balancing accuracy and efficiency. Likewise, Fang et al.\cite{fang2024pest} proposed Pest-ConFormer, a CNN-transformer Hybrid with multi-scale feature selection, which achieved state-of-the-art results (accuracy 77.81\%) on a large-scale multi-class IP102 benchmark dataset.

Overall, hybrid architectures underscore the complementary nature of CNNs and transformers. CNN components excel at recognizing texture or fine details like wing venation or spot patterns, whereas transformer components capture the broader context such as pest position on a leaf or co-occurrence of multiple pests in one image. The synergy of these aspects leads to more robust models, particularly valuable in the complex scenes of agricultural settings. Table \ref{tab:hybrid-litreview} lists an overview of recent hybrid-CNN approaches for crop pest and disease detection.

\begin{table}[!ht]
\centering
\caption{
\textbf{Literature Review Table on Hybrid CNN-Transformer-Based Approaches.}
}
\tiny
\renewcommand{\arraystretch}{2}
\begin{tabular}{p{0.5cm} p{0.5cm} p{14cm}}
\hline
\textbf{Ref} & \textbf{Year} & \textbf{Description} \\[1.2ex]
\hline
\cite{utku2025convvit} & 2025 & Introduced ConvViT for insect detection, a CNN+ViT hybrid optimized model, combining high accuracy with low model size. \\[1.2ex]\hline
\cite{hu2024tea} & 2024 & Proposed TP-YOLOX for tea pests in complex backgrounds, a module combining convolution with multi-head self-attention plus a multi-scale attention module. +4.50\% mAP over YOLOX-s at 82.66 FPS. \\[1.2ex]\hline
\cite{fang2024pest} & 2024 & Proposed Pest-ConFormer, a multi-scale hybrid CNN-transformer architecture for large-scale pest recognition, achieving state-of-the-art results. \\[1.2ex]\hline
\end{tabular}
\begin{flushleft}
\footnotesize
\end{flushleft}
\label{tab:hybrid-litreview}
\end{table}

\vspace{-2em}

\subsection{Object Detection Models}\label{sec:object_detection}
While classification (assigning a single label to an image) is useful, practical field scenarios often involve multiple pests per image or a need to pinpoint pest locations. Thus, object detection models have been applied to agricultural pest monitoring, with frameworks like Faster R-CNN and especially YOLO becoming dominant. In recent years, YOLO-based approaches have been favored for their real-time performance and high accuracy trade-off. Amin et al.\cite{amin2} utilized YOLOv5 to localize pests in field images, illustrating how detection networks can both identify and locate insects within an image frame. Building on this, Du et al.\cite{Du2024} and Wang et al.\cite{Wang2024micro} experimented with newer YOLO versions (YOLOv7, YOLOv8) and reported improved detection accuracy of 98.16\% and speed thanks to architectural advancements in those versions. For instance, N. Wang et al.\cite{wang2025insect} (the Insect-YOLO study discussed below) also performed a regression analysis on the detection outputs to estimate pest counts and reported an $R^2$ of 0.99 against manual counting. Such a high correlation suggests that detection models can support quantitative pest monitoring, a key requirement for Integrated Pest Management (IPM) programs. On the other hand, Faster R-CNN has served as a strong two-stage detector baseline. The systematic review by Teixeira et al.\cite{teixeira2023systematic} found that the best-performing detector differed between studies, as some of the compared studies reported YOLO-family detectors as the best performers and others reported Faster R-CNN, and the review's top-ranked detection results were a modified Faster R-CNN (95.20\% mAP), YOLOv5 (94.70\%) and Faster R-CNN (94.64\%). Performance therefore depends on the model variant, the dataset, and the evaluation protocol. Even so, recent work in pest detection frequently favors YOLO variants or similar one-stage architectures because of their real-time speed and ease of adaptation. Researchers have introduced domain-specific enhancements to these models. N. Wang et al.\cite{wang2025insect} developed an Insect-YOLO model tailored to capturing small pests on various crops, featuring streamlined parameters for speed and custom data augmentation for diversity. This specialization yielded exceptional accuracy with real-time speeds, addressing some challenges of deploying detectors in the field. Similarly, Yu and Zhang \cite{yu2023mdpyolo} proposed MDP-YOLO, a lightweight YOLOv5s-based detector for rice pests that adds a large-scale feature extraction layer to better detect tiny insects, increasing mean average precision by 5.5\% over the original YOLOv5s. These targeted improvements show that with careful adaptation, general-purpose models like YOLO can be tuned to agricultural needs, e.g., focusing on small object detection or handling dense pest clusters on traps. An important aspect of detection in agriculture is dealing with class imbalance (many pest instances might belong to a few common species, with others being rare). Researchers often incorporate techniques like hard example mining or focal loss into detectors to bias training towards those hard-to-detect, less frequent pests. For instance, Du et al.\cite{Du2024} emphasized training on hard examples (e.g., pests camouflaged against leaves) to reduce background false positives. Such strategies improved robustness in complex field imagery. In summary, object detection models, particularly the YOLO family, are crucial for advancing from simply saying “this image has pest X” to answering “where and how many pests are present.” They enable automated pest counting and mapping, which are invaluable for precision agriculture. 

The literature trend clearly shows YOLO-based methods leading the pack for pest detection, owing to their speed and adaptability, with continual refinements to address domain-specific challenges like tiny object size and cluttered backgrounds. Table \ref{tab:object-detect-litreview} summarizes the recent object detection models for agricultural pest detection and monitoring.

\begin{table}[!ht]
\centering
\caption{
\textbf{Literature Review Table on Object Detection Models.}
}
\tiny
\renewcommand{\arraystretch}{2}
\begin{tabular}{p{0.5cm} p{0.5cm} p{14cm}}
\hline
\textbf{Ref} & \textbf{Year} & \textbf{Description} \\[1.2ex]
\hline
\cite{teixeira2023systematic} & 2023 & Systematic review of deep learning for insect detection (trap images) found no single best detector as its top reported results were a modified Faster R-CNN (95.20\% mAP), YOLOv5 (94.70\%) and Faster R-CNN (94.64\%), and performance depended on the model variant and dataset. \\[1.2ex]\hline
\cite{wang1122} & 2025 & Proposed RGC-YOLO, a lightweight YOLOv8n-based detector (RepGhost reparameterization, GhostConv, CBAM attention) for multi-scale rice pest and disease detection, reached 93.2\% mAP50 and was deployed on a Jetson Nano embedded device. \\[1.2ex]\hline
\cite{amin2} & 2023 & Employed YOLOv5 for pest localization in field images, demonstrating that one-stage detectors can accurately detect multiple pests per image. \\[1.2ex]\hline
\cite{Du2024} & 2025 & Applied YOLOv7 to pest detection with improved accuracy, utilizing hard example mining to better learn from challenging, camouflaged pest instances. \\[1.2ex]\hline
\cite{wang2025insect} & 2025 & Created “Insect-YOLO” with custom enhancements for multi-crop pest detection (streamlined parameters, specialized augmentations), achieving real-time detection with high accuracy and reporting a high correlation ($R^2 = 0.99$) between automated and manual pest counts. \\[1.2ex]\hline
\cite{yu2023mdpyolo} & 2023 & Proposed MDP-YOLO, a lightweight YOLOv5s-based detector for rice pests (ShuffleNetV2 backbone) and added a large-scale feature extraction layer to improve detection of small pests, increasing mAP by 5.5\% over YOLOv5s. \\[1.2ex]\hline
\end{tabular}
\begin{flushleft}
\footnotesize
\end{flushleft}
\label{tab:object-detect-litreview}
\end{table}

\subsection{Emerging and Non-Traditional AI Techniques}
Beyond mainstream deep learning approaches, a small but growing number of studies are exploring novel AI paradigms. Amin et al.\cite{amin2} ventured into quantum machine learning, applying a quantum CNN to a pest classification task. Although the model was only tested on a small dataset, it achieved near-perfect accuracy, showcasing intriguing early potential for quantum-enhanced image processing, albeit in a nascent, experimental stage. More practical in the short term are self-supervised learning methods. Kar et al.\cite{Kar} utilized BYOL (Bootstrap Your Own Latent) to pre-train pest classification models using unlabeled data. Their approach nearly matched the performance of fully supervised counterparts, highlighting a valuable strategy for overcoming the labeling bottleneck in agricultural datasets. Additionally, interdisciplinary methods such as metaheuristic optimization have been applied. For instance, Moth Flame Optimization (MFO) was used by Venkatasaichandrakanth and Iyapparaja \cite{Venkatasaichandrakanth} to enhance feature selection in their EViTA-based peanut pest classifier. These methods, while less conventional in computer vision, demonstrate potential for improving performance, particularly in small-sample or constrained-data settings. Table \ref{tab:emerging-ai-litreview} summarizes the novel and non-traditional AI methods recently explored for crop pest detection and decision support. The technical landscape of AI in pest classification is rapidly evolving from foundational CNN classifiers to transformer-centric and hybrid architectures, and even experimental paradigms like quantum learning and self-supervised approaches.

\begin{table}[!ht]
\centering
\caption{
\textbf{Literature Review Table on Emerging and Non-Traditional Approaches.}
}
\tiny
\renewcommand{\arraystretch}{2}
\begin{tabular}{p{0.5cm} p{0.5cm} p{14cm}}
\hline
\textbf{Ref} & \textbf{Year} & \textbf{Description} \\[1.2ex]
\hline
\cite{amin2} & 2023 & Explored a quantum convolutional neural network for pest classification, achieving near-perfect accuracy on a small dataset, showing early promise for quantum-enhanced learning. \\[1.2ex]\hline
\cite{du2025longtail} & 2025 & Applied a diffusion model to generate synthetic pest images for data augmentation, improving long-tail pest class accuracy significantly by augmenting scarce classes. \\[1.2ex]\hline
\cite{Venkatasaichandrakanth} & 2023 & Used Moth Flame Optimization (MFO), a metaheuristic, for feature selection in the EViTA peanut-pest classifier. \\[1.2ex]\hline
\cite{Kar} & 2023 & Used self-supervised BYOL to pre-train on unlabeled pest images; the fine-tuned model reached $\sim$94\% accuracy on 12-class pest identification, nearly matching fully supervised results. \\[1.2ex]\hline
\cite{karam2022gan} & 2022 & Developed a GAN-based augmentation tool for pest detection (whiteflies on traps), and augmented data led to higher detection rates, with recall from 54.4\% to 94.6\% for rare pest instances. \\[1.2ex]\hline
\cite{soni2022pest} & 2022 & Integrated pest CNN models with an Augmented Reality app for farmers, enabling real-time pest identification through a smartphone camera with overlaid guidance (enhancing practical usability). \\[1.2ex]\hline
\end{tabular}
\begin{flushleft}
\footnotesize
\end{flushleft}
\label{tab:emerging-ai-litreview}
\end{table}

\begin{figure}[ht!]
\centering
\includegraphics[width=\textwidth]{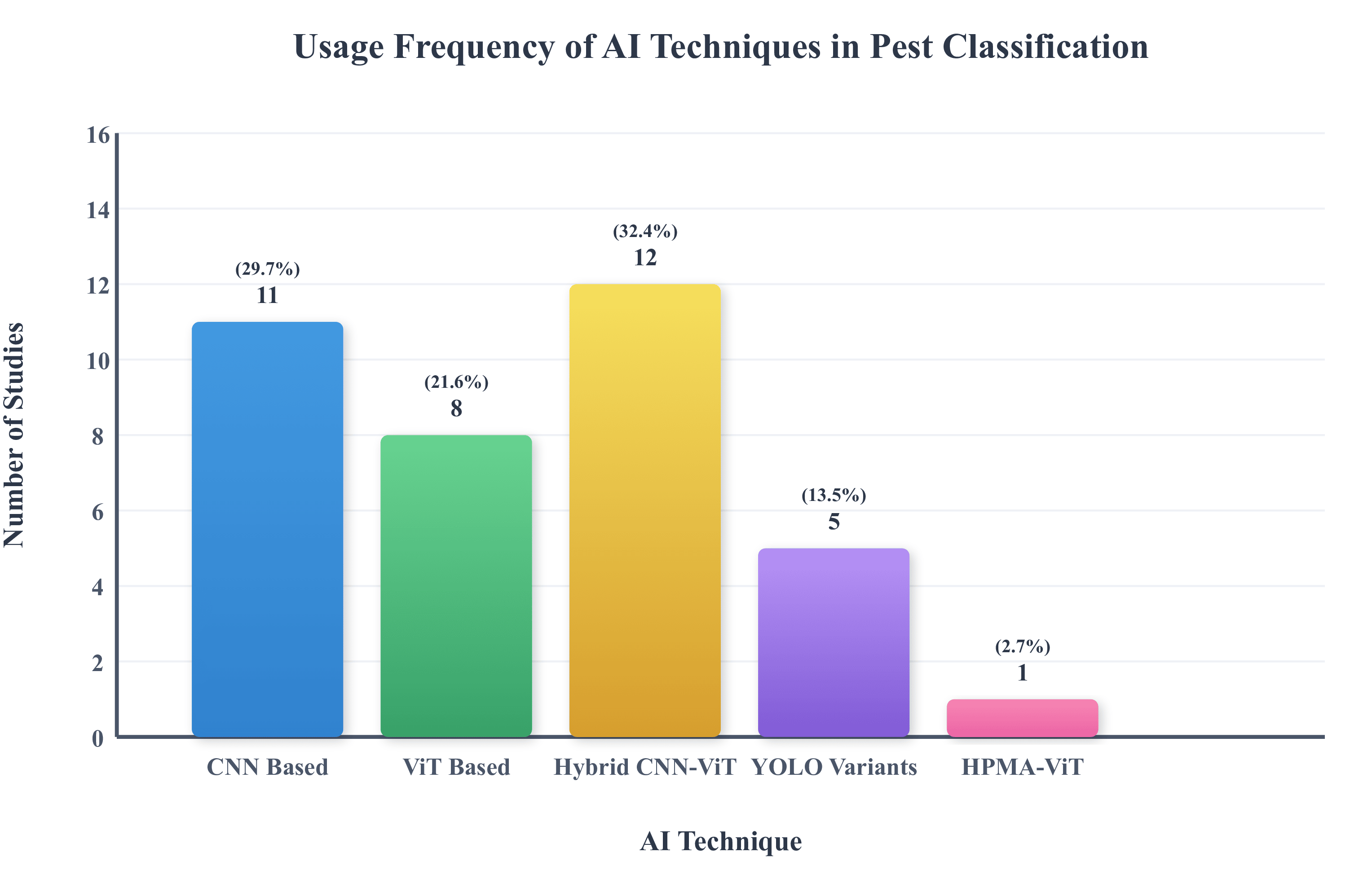}
\caption{\textbf{Usage Frequency of AI Techniques in Pest Classification}}
\label{fig:AITechq}
\end{figure}

The literature consistently shows that using advanced architectures such as ViTs, attention modules, or hybrid combinations yields significant performance improvements. However, such gains are often balanced against increased model complexity and resource demands. As a result, there is growing interest in designing lightweight yet effective models. For example, Xiao et al.\cite{xiao2024ga} and Wang et al.\cite{wang1122} focused on efficiency through model compression or architectural streamlining to enable on-device deployment, an important consideration for field applications where compute resources may be limited. Ultimately, the diversity of approaches provides researchers and practitioners with a robust toolkit. The choice of model architecture and training strategy will depend heavily on specific use-case requirements, such as the need for speed, interpretability, dataset size, or deployment environment. This rich ecosystem of AI techniques underscores an exciting frontier in pest management, where technology is increasingly capable of providing timely, accurate, and scalable solutions. Figure \ref{fig:AITechq} demonstrates the usage frequency of AI techniques in pest classification. 

\section{Benchmark Datasets}\label{sec:datasets}
A variety of image datasets have been used in the literature for training and evaluating pest classification models. Here we describe the most prominent datasets, including their content (number of classes, images), availability, and other attributes like resolution or source, as these factors greatly influence model development and performance.

\subsection{IP-102}
IP102 - Insect Pest 102 dataset (Wu et al.\cite{wu2019ip102}). This is one of the largest public datasets for pest recognition. It contains 102 insect pest classes across about 75,000 images. The images are collected from a variety of crops (around 8 major crop types) and include both field photographs and some lab images. The average number of images per class is 737, but the distribution is highly imbalanced. Image resolutions vary, but many are medium resolution (e.g., 500×500 pixels or higher) in color. IP102 is publicly available (hosted with the original paper) and has become a benchmark. For example, baseline CNN accuracy was 49, highlighting its difficulty. This dataset covers a broad range of pest types (chewers, sap-suckers, borers, etc.) and is valuable for evaluating general-purpose
models.
\subsection{Xei-1}
Xie-1 (2015) refers to the dataset used by Xie et al.\cite{xie2015multitask}, which covers 24 common field crop insect species. It is reported to contain about 1,440 images, which would be quite limited for CNN training. Due to its small size, Xie1 alone is not sufficient for deep learning, but it has been used in combination with other data or for transfer learning purposes.

\subsection{Xei-2}
Xie et al.\cite{xie2018multilevel} and collaborators expanded on earlier datasets. Their study describes a dataset of 40 insect classes with about 4,500 images. This is also known as the D0 dataset. It includes more classes than the previous version. Xie-2 serves as an intermediate-scale dataset and has been used to pre-train models or evaluate alongside IP102. Its imbalance is less extreme than IP102 but still present.

\subsection{Plant Village}
PlantVillage \cite{plantvillagehughes} is a well-known open dataset of plant disease images, with over 50,000 curated images of healthy and infected leaves of various crops. The widely used release analysed by Mohanty et al.\cite{mohanty2} contains 54,306 images in 38 crop-disease classes. It is a plant-disease dataset with classes of crop-disease pairs. Li et al.\cite{li2022plant} worked with a 39-class dataset of plant disease and insect pest images because the original PlantVillage classes are crop-disease pairs. Images are high-quality, mostly lab-controlled (solid background with a single leaf). Resolution is high (many 256×256 or 512×512 after preprocessing). PlantVillage data tends to yield very high model accuracies (due to uniform images), but models trained solely on it might not generalize to field conditions. Nonetheless, it is a useful training resource for plant-health recognition and is often used for transfer learning.

\begin{table*}[!ht]
\centering
\caption{
\textbf{Summary of Notable Pest Image Datasets.}
}
\tiny
\renewcommand{\arraystretch}{2}
\begin{tabular}{p{0.5cm} p{1cm} p{2cm} p{2cm} p{0.5cm} p{3.5cm} p{4cm}}
\hline
\textbf{Ref} & \textbf{Dataset (Year)} & \textbf{Classes} & \textbf{Images} & \textbf{Public} & \textbf{Typical Image Resolution} & \textbf{Notes} \\[1.2ex]
\hline
\cite{alshannaq} & Tomato Pest (2025) & 8 (insect-pest categories) & 842 (raw) / 5,894 (aug) & No & 1080$\times$1080 px (smartphone) & Augmented for training with eight-category classification. \\[1.2ex]\hline
\cite{berka2023cactivit} & CactiViT (2023) & 3 (infestation levels) & 3921 & Yes & 720$\times$1280 px (smartphone) & Open-source domain specific dataset. \\[1.2ex]\hline
\cite{wu2019ip102} & IP102 (2019) & 102 & 75,000 & Yes & 500$\times$500 px (varied field shots) & Benchmark for fine-grained pest recognition. \\[1.2ex]\hline
\cite{amin2} & Kaggle Paddy Pest (2022) & 2 (pest vs none) & 648 & Yes & 640$\times$480 px (varied) & Useful for simple detection but highly unbalanced. \\[1.2ex]\hline
\cite{Du2024} & Corn-Rice Pest (2025) & 5 & 4620 & No & 640$\times$640 px (resized) & Rice-corn pest dataset used for YOLOv7-PSAFP. \\[1.2ex]\hline
\cite{liu2016paddy} & Rice Pest 12 (2016) & 12 & $>$5,000 & No & 256$\times$256 px (cropped, resized) & Rice field pest images, moderately sized and used in early CNN work. \\[1.2ex]\hline
\cite{LIU2025109920} & Forestry Pest (2025) & 31 main + 16 sub & 7163 & Yes & 224$\times$224 px (resized) & Forestry pest dataset used with a transformer-based detector. \\[1.2ex]\hline
\cite{xie2015multitask} & Xie1 (2015) & 24 & $\sim$1,440 & No & n/r & Early dataset of common field-crop insects; reportedly small per class ($\sim$60). \\[1.2ex]\hline
\cite{xie2018multilevel} & Xie2/D0 (2018) & 40 & $\sim$4,500 & No & n/r & Includes more species than Xie1 (field crop insects). \\[1.2ex]\hline
\cite{plantvillagehughes} & PlantVillage (2015) & 38 (crop-disease pairs) & 54,306 & Yes & 256$\times$256 px (processed) & Plant-disease dataset, lab-like images, high quality. \\[1.2ex]\hline
\cite{petchiammal2023paddy} & Paddy Doctor (2023) & 13 (12 disease/pest + normal) & 16,225 & Yes & 1080$\times$1440 px (smartphone) & Field-captured paddy leaf images, most classes are diseases, but five are insect-pest classes (hispa, leaf roller and 3 types of stem borers). \\[1.2ex]\hline
\end{tabular}
\begin{flushleft}
\footnotesize
\textit{Note:} n/r = not reported in the sources consulted. $\sim$ = approximate.
\end{flushleft}
\label{tab:datasets-pest}
\end{table*}

\vspace{-2em}

\subsection{Other Datasets}
In addition to established benchmark datasets like IP102 and PlantVillage, many studies rely on specialized or custom-collected datasets designed for specific pests, crops, or regional field conditions. These datasets, while often limited in size or public accessibility, provide high contextual relevance and are instrumental in developing real-world-ready AI models. For instance, Al-Shannaq et al.\cite{alshannaq} developed a tomato pest dataset containing 842 field-captured images, augmented to 5,894 samples, which was used to train a CNN to classify eight categories of tomato insect pests. Berka et al.\cite{berka2023cactivit} introduced the CactiViT dataset with 3921 annotated images of cochineal infestations on cacti with varying resolutions and released it as an open-source resource. Du et al.\cite{Du2024} evaluated their detector on a rice-corn pest dataset (corn and rice pests such as borers and planthoppers) with 4620 samples. Additional smaller-scale public datasets have also been employed. Amin et al.\cite{amin2} used the Kaggle Paddy Pest Dataset, which includes 648 images (135 pest-positive and 513 pest-free), for binary pest detection using hybrid quantum CNNs. The Rice Pest dataset of Liu et al.\cite{liu2016paddy}, containing over 5,000 images of 12 rice pest classes, was integral to early CNN-based pest classification. Another recent addition is the forestry pest dataset used by Liu et al.\cite{LIU2025109920}, on which their transformer-based detector reportedly achieved over 95\% mAP, divided into 31 main categories and 16 subcategories with 7163 publicly available images. These datasets collectively highlight the importance of localized, context-specific data in advancing the effectiveness and deployment of AI systems in agricultural pest management. Table \ref{tab:datasets-pest} covers a range of crops, pest species, and image acquisition conditions. Public availability and contextual fit are key to model generalizability.

\section{Challenges}\label{sec:challenges_future}
\subsection{Dataset Imbalance}
A foundational challenge in pest classification lies in the issue of imbalanced and insufficient datasets. Most existing datasets show a stark imbalance in class distribution: common pests tend to dominate with thousands of images, while rare or region-specific pests may have only a handful of annotated samples. This imbalance can significantly skew model performance, often resulting in high accuracy for dominant classes and poor generalization for underrepresented ones. Fu et al.\cite{fu2024} directly noted this issue, reporting reduced model accuracy for classes with fewer samples. To counteract this, researchers have adopted a range of strategies. Data augmentation is ubiquitous, used to synthetically expand dataset diversity. Loss function modifications like focal loss, which Venkatasaichandrakanth and Iyapparaja describe in GNViT as a remedy for class imbalance \cite{venkatasaichandrakanth2024gnvit} (although the same paper also refers to a cross-entropy loss, so the loss actually used is not described consistently), aim to reduce the bias towards frequent classes. However, these methods offer only partial solutions. The deeper issue is the limited availability of comprehensive, high-quality pest image datasets. Al-Shannaq et al.\cite{alshannaq} and Lin et al.\cite{lin2023GPAN} both identified the scarcity of annotated images as a critical bottleneck for robust AI development. Kar et al.\cite{Kar} demonstrated how self-supervised learning, particularly the BYOL approach, could leverage large volumes of unlabeled data to pre-train models that approach supervised performance. Moving forward, promising directions include the creation of open-access, crowd-sourced pest image repositories and the use of few-shot and semi-supervised learning techniques. These approaches could help democratize data availability, enabling more inclusive and generalized AI systems for pest classification. Table \ref{tab:longtail-datascarcity} highlights the challenges and strategies for handling class imbalance and data scarcity in agricultural pest recognition.

\begin{table}[!ht]
\centering
\caption{
\textbf{Key Studies Addressing Data Scarcity and Long-Tail Distribution in Pest Recognition.}
}
\tiny
\renewcommand{\arraystretch}{2}
\begin{tabular}{p{0.5cm} p{0.5cm} p{14cm}}
\hline
\textbf{Ref} & \textbf{Year} & \textbf{Description} \\[1.2ex]
\hline
\cite{alshannaq}& 2025 & Emphasized the critical shortage of localized pest image datasets (e.g., for Jordan crops), identifying data scarcity as a key limitation. \\[1.2ex]\hline
\cite{wu2019ip102} & 2019 & Released IP102, a 75,000+ image dataset of 102 pest species with a naturally imbalanced distribution, a benchmark that spurred research into long-tail learning for pests. \\[1.2ex]\hline
\cite{venkatasaichandrakanth2024gnvit} & 2024 & Describes focal loss as a remedy for class imbalance in a four-class groundnut pest dataset, but also refers to a cross-entropy loss, so the loss actually implemented is not described consistently. \\[1.2ex]\hline
\cite{Kar}& 2023 & Demonstrated self-supervised pre-training (BYOL) on unlabeled pest images to combat data scarcity and achieved near supervised-level accuracy with far fewer labeled examples. \\[1.2ex]\hline
\cite{lin2023GPAN} & 2023 & Similarly pointed out the lack of comprehensive, annotated pest image repositories and called for community efforts in data collection and sharing. \\[1.2ex]\hline
\cite{fu2024} & 2024 & Noted significant accuracy drop on under-represented pest classes. Highlighted how long-tailed training data skewed ViT performance. \\[1.2ex]\hline
\cite{wang2021agripest} & 2021 & Created AgriPest, a large-scale pest dataset (49.7k images, 14 species) captured in real field conditions and aimed to provide a realistic benchmark for training and evaluation. \\[1.2ex]\hline
\end{tabular}
\begin{flushleft}
\footnotesize

\end{flushleft}
\label{tab:longtail-datascarcity}
\end{table}

\vspace{-2em}

\subsection{Complex Field Environments}
Another major hurdle is the variability and complexity of real-world field conditions. Unlike controlled laboratory environments, crop fields present cluttered backgrounds composed of leaves, stems, soil, and debris, combined with inconsistent lighting conditions (e.g., glare, shadows, low-light scenarios). These elements introduce noise and visual distractions that significantly challenge model robustness and generalization. This challenge has been widely acknowledged in the literature. Liu et al.\cite{LIU2025109920} introduced a regional pale-shaped self-attention (RPSA) mechanism in their transformer-based detector to account for background context, reporting improved detection performance in visually busy field images. Du et al.\cite{Du2024} tackled the issue by emphasizing hard example mining during training, which helped reduce false positives arising from background confusion. Despite these advancements, the problem is far from solved. In the tea-pest study of Zejun Wang et al.\cite{Wang2024micro}, the detection confidence of all compared models dropped to some extent on external images taken under cloudy, rainy and low-light conditions, especially when variable light intensity coincided with incomplete pest bodies, even though brightness augmentation had been used in training. Such findings point to a need for improved data augmentation strategies that simulate various lighting and environmental conditions. Additionally, domain adaptation techniques, where models are trained across diverse domains or geographical settings, could improve generalization and robustness. Future work must continue to address the environmental variability inherent in real-world deployment scenarios. Table \ref{tab:field-complexity-robustness} highlights the Studies Tackling Field Complexity and Robustness in agricultural pest recognition.

\begin{table}[!ht]
\centering
\caption{
\textbf{Selected Studies Tackling Field Complexity and Robustness in Pest Detection.}
}
\tiny
\renewcommand{\arraystretch}{2}
\begin{tabular}{p{0.5cm} p{0.5cm} p{14cm}}
\hline
\textbf{Ref} & \textbf{Year} & \textbf{Description} \\[1.2ex]
\hline
\cite{Du2024} & 2025 & Employed hard example mining during training to teach the model using cluttered, difficult images and reduced false positives from leaves and background noise. \\[1.2ex]\hline
\cite{wang2025insect} & 2025 & Applied data augmentation (e.g., cropping, rotation, and flipping) to diversify the training images of its crop-insect detector, a common first step toward robustness to varying backgrounds and lighting. \\[1.2ex]\hline
\cite{Wang2024micro} & 2024 & On external images taken under cloudy, rainy and low-light conditions, the detection confidence of all compared models decreased to some extent, especially when variable light intensity coincided with incomplete pest body brightness augmentation used in training to simulate varied lighting. \\[1.2ex]\hline
\cite{LIU2025109920} & 2025 & Introduced regional pale-shaped self-attention (RPSA) in their transformer-based detector to handle cluttered backgrounds and improve pest detection in visually busy field images by focusing on relevant regions. \\[1.2ex]\hline
\end{tabular}
\begin{flushleft}
\footnotesize
\end{flushleft}
\label{tab:field-complexity-robustness}
\end{table}

\vspace{-2em}

\subsection{Detection of Small Pests}
A significant technical difficulty lies in the detection of small pests, which often occupy only a tiny portion of the image. This creates problems for both classification and detection models, as the visual features of small objects are often lost in deep layers of neural networks. It’s a known limitation even in general computer vision, and it’s especially acute in agricultural applications where pests can be millimeters in size. Several researchers have proposed model-specific solutions. Hu et al.\cite{hu2023detection} added an additional fine-scale detection head to YOLO to capture small pest signatures more effectively. Wang et al.\cite{wang1122} incorporated Convolutional Block Attention Module (CBAM) attention into their detection pipeline, enhancing sensitivity to small, localized objects. Likewise, the systematic review by Teixeira et al.\cite{teixeira2023systematic} identified small object detection as a consistent challenge across detection benchmarks. Potential future directions include leveraging super-resolution techniques to enhance small objects before detection or combining traditional RGB imaging with specialized sensors (macro lenses, hyperspectral cameras). Additionally, techniques like multi-scale feature pyramids and transformer-based aggregation, such as in Liu et al.\cite{LIU2025109920}, have proven valuable in maintaining detailed spatial information. At the hardware level, deploying high-resolution, close-range cameras such as smart traps or crop-mounted sensors can support better detection, though they require corresponding algorithmic adjustments to manage the increased data load. The above-mentioned studies in Table \ref{tab:small-pest-detection} demonstrate state-of-the-art techniques for improving the detection and classification of tiny insects in field and trap-based settings.

\begin{table}[!ht]
\centering
\caption{
\textbf{Selected Studies Targeting Small Pest Detection in Agricultural Images.}
}
\tiny
\renewcommand{\arraystretch}{2}
\begin{tabular}{p{0.5cm} p{0.5cm} p{14cm}}
\hline
\textbf{Ref} & \textbf{Year} & \textbf{Description} \\[1.2ex]
\hline
\cite{teixeira2023systematic} & 2023 & Identified small object detection as a major challenge across studies, noting that models perform significantly worse on insects that occupy few pixels, urging focus on multi-scale methods. \\[1.2ex]\hline
\cite{wang1122} & 2025 & Incorporated CBAM attention into the detection pipeline, heightening the model's sensitivity to small, localized pest features. \\[1.2ex]\hline
\cite{hu2023detection} & 2023 & Added an extra fine-scale detection head in their YOLO-based model to expand the detection scale range, improving detection of various-sized insects that standard heads often miss. \\[1.2ex]\hline
\cite{wang2024deep} & 2024 & Developed a method for detecting 3 insects, including whiteflies on yellow sticky traps (small targets) using high-resolution input and an improved YOLOv7, achieving high mean average precision on 11 compared baselines. \\[1.2ex]\hline
\cite{yang2024srnetyolo} & 2024 & Proposed SRNet-YOLO for tiny and very tiny pests in cotton fields, a feature-map super-resolution (FM-SR) module inside a YOLOv8-based detector reconstructs deep feature maps at a higher spatial resolution before feature fusion, and reported 78.2\% mAP for tiny and 57\% mAP for very tiny-pest detection. \\[1.2ex]\hline
\end{tabular}
\begin{flushleft}
\footnotesize
\end{flushleft}
\label{tab:small-pest-detection}
\end{table}

\subsection{Computational Cost}
Beyond accuracy, the real-world deployment of pest detection models demands computational efficiency and responsiveness. Several studies have highlighted the tension between model complexity and deployability. High-capacity models like Vision Transformers offer superior accuracy but require considerable resources, making them impractical for edge devices or mobile applications commonly used in agricultural fields.

Al-Shannaq et al.\cite{alshannaq} pointed to mobile applications as a potential use of their model, presented as a possible application rather than a demonstrated on-device deployment. Berka et al.\cite{berka2023cactivit} successfully deployed a pest classification system on a mobile device, showing that real-time applications are feasible but require careful model design. To reduce computational demand, several strategies are being explored. Xiao et al.\cite{xiao2024ga} used GA-GhostNet, a streamlined CNN architecture that balances efficiency and accuracy by significantly reducing model parameters. Other potential approaches include pruning, quantization, and knowledge distillation, where a small “student” model is trained to mimic the behavior of a large, accurate “teacher” model. Although knowledge distillation remains underexplored in pest classification, it holds promise as a technique to deliver high performance in resource-constrained environments. Table \ref{tab:mobile-pest-ai} highlights the studies' progress and innovation in real-time, resource-efficient pest identification for mobile and edge computing.

\begin{table}[!ht]
\centering
\caption{
\textbf{Recent Studies on Mobile/Edge Deployment for Pest Detection.}
}
\tiny
\renewcommand{\arraystretch}{2}
\begin{tabular}{p{0.5cm} p{0.5cm} p{14cm}}
\hline
\textbf{Ref} & \textbf{Year} & \textbf{Description} \\[1.2ex]
\hline
\cite{alshannaq} & 2025 & Suggested that the trained CNN could be used in mobile applications to help farmers detect pests as a potential application rather than a demonstrated on-device deployment. \\[1.2ex]\hline
\cite{berka2023cactivit} & 2023 & Demonstrated a successful on-device pest classification app that used model optimization to achieve real-time inference on mobile hardware. \\[1.2ex]\hline
\cite{xiao2024ga} & 2024 & Utilized GA-GhostNet (an efficient CNN) to balance accuracy and speed, significantly reducing model parameters while maintaining high precision. \\[1.2ex]\hline
\cite{cheng2022lightweight} & 2022 & Proposed YOLOLite-CSG, a lightweight detector derived from a simplified YOLOv3 (about 5 M parameters, 9.8 GFLOPs), reached 82.9\% precision on the CP15 crop pest dataset, illustrating a strong accuracy-efficiency trade-off for field use. \\[1.2ex]\hline
\cite{jiang2024forest} & 2024 & Improved YOLOv8n for forest pest recognition (7 pest classes) with a GSConv Slim-Neck, CBAM attention and WIoU loss achieved 98.9\% precision and 97.6\% recall in a lightweight model. \\[1.2ex]\hline
\cite{hinton2015distilling} & 2015 & General machine-learning technique, not a pest study, knowledge distillation, in which a smaller model learns from a larger model's outputs, suggested as a possible future route to compact pest classifiers. \\[1.2ex]\hline
\end{tabular}
\begin{flushleft}
\footnotesize
\end{flushleft}
\label{tab:mobile-pest-ai}
\end{table}

\vspace{-2em}

\subsection{Limited Generalization}
Another open issue is the limited generalization capacity of current models. Many are trained and evaluated on narrowly defined datasets, often from a single region, crop type, or image source. While they perform well within these boundaries, their accuracy can degrade sharply when applied to new environments, crops, or pest varieties. This poses a major barrier to the widespread adoption of AI in diverse agricultural settings. Some progress has been made. Xiao et al.\cite{xiao2024ga} explored cross-dataset transfer, showing that a model trained on pest images could be fine-tuned for disease recognition. However, transfer learning across different pest datasets or geographic regions is still largely under-investigated. GA-GhostNet’s limitations in generalization further underscore this gap. Lin et al.\cite{lin2023GPAN} discussed extending their approach to broader agricultural applications, including temporal weather prediction, grain storage management, and Internet of Medical Things (IoMT) systems. A future-ready pest detection system must be able to handle new classes and unforeseen pest appearances. This calls for approaches like open-set recognition, continual learning, and active learning. In particular, models that can flag uncertain predictions for human labeling and then incorporate new knowledge over time would significantly improve adaptability. This is increasingly important as climate change alters pest distributions, introducing novel species into previously unaffected areas. Therefore, model adaptability and long-term learning capabilities will be essential for sustainable pest management in evolving agricultural ecosystems. The studies mentioned in Table \ref{tab:ai-generalization} illustrate the current best practices and emerging strategies for robust, adaptable pest and plant disease AI across varied and evolving agricultural contexts.

\begin{table}[!ht]
\centering
\caption{
\textbf{Recent Advances and Challenges in Generalization, Transfer Learning, and Adaptability for Pest and Plant Disease AI.}
}
\tiny
\renewcommand{\arraystretch}{2.5}
\begin{tabular}{p{0.5cm} p{0.5cm} p{14cm}}
\hline
\textbf{Ref} & \textbf{Year} & \textbf{Description} \\[1.2ex]
\hline
\cite{xiao2024ga} & 2024 & Found that models trained on one pest dataset did not generalize well to another; cross-dataset transfer learning required fine-tuning, indicating domain-specific bias. \\[1.2ex]\hline
\cite{lin2023GPAN} & 2023 & Highlighted the need for broader generalization, suggesting integrating environmental data for predictive modeling, but noted this remains speculative without further research. \\[1.2ex]\hline
\cite{parisi2019continual} & 2019 & Reviewed continual learning techniques relevant to AI for updating models with new classes over time without forgetting old ones, thereby enhancing long-term generalization. \\[1.2ex]\hline
\cite{yang2024fewshot} & 2024 & Applied cross-domain few-shot learning to plant disease identification, letting pest classifiers learn new categories from only a few examples. \\[1.2ex]\hline
\end{tabular}
\begin{flushleft}
\footnotesize
\end{flushleft}
\label{tab:ai-generalization}
\end{table}

\section{Conclusion}\label{sec:practical_applications}
This review discusses how pest classification has quickly developed with deep learning, from conventional CNNs to stronger hybrid and transformer-based models that yield superior accuracy and versatility in agricultural settings. After reviewing 37 studies, it is apparent that the majority of work involves crop plants such as rice, groundnut, and tomato, with emerging interest in others like cactus and tea. Whereas image-based approaches utilizing platforms such as CCTV and drones are prominent, sensor-based and smartphone-friendly solutions are making inroads for real-time application in the field. Nonetheless, significant bottlenecks exist, such as skewed datasets, difficulty seeing small or concealed pests, and models that don’t generalize well between regions or crops. There are promising new improvements in quantum-inspired CNNs, data augmentation techniques such as diffusion models, and edge-friendly versions of YOLO. Future work must address building larger and more diverse open datasets, developing energy-efficient models that can be deployed in the field, and building systems that generalize well under varying agricultural conditions. Transitioning research to farms will necessitate strong collaboration among AI researchers and agronomists.

\bibliography{ref}

\end{document}